# Ensembles of Kernel Predictors

**Corinna Cortes**
Google Research
76 Ninth Avenue
New York, NY 10011
corinna@google.com

**Mehryar Mohri**
Courant Institute and Google
251 Mercer Street
New York, NY 10012
mohri@cims.nyu.edu

**Afshin Rostamizadeh**
UC Berkeley
Sutardja Dai Hall
Berkeley, CA 94720
arostami@eecs.berkeley.edu

## Abstract

This paper examines the problem of learning with a finite and possibly large set of $p$ base kernels. It presents a theoretical and empirical analysis of an approach addressing this problem based on ensembles of kernel predictors. This includes novel theoretical guarantees based on the Rademacher complexity of the corresponding hypothesis sets, the introduction and analysis of a learning algorithm based on these hypothesis sets, and a series of experiments using ensembles of kernel predictors with several data sets. Both convex combinations of kernel-based hypotheses and more general $L_q$-regularized non-negative combinations are analyzed. These theoretical, algorithmic, and empirical results are compared with those achieved by using learning kernel techniques, which can be viewed as another approach for solving the same problem.

## 1 Introduction

Kernel methods are used in a variety of applications in machine learning [22]. Positive definite (PDS) kernels provide a flexible method for implicitly defining features in a high-dimensional space where they represent an inner product. They can be combined with large-margin maximization algorithms such as support vector machines (SVMs) [8] to create effective prediction techniques.

The choice of the kernel is critical to the success of these algorithms, thus committing to a single kernel could be suboptimal. It could be advantageous instead to specify a finite and possibly large set of $p$ base kernels. This leads to the following general problem central to this work: ($\mathcal{P}$) how can we best learn an accurate predictor when using $p$ base kernels?

One approach to this problem is known as that of *learning kernels* or *multiple kernel learning* and has been extensively investigated over the last decade by both algorithmic and theoretical studies [16, 2, 1, 23, 17, 26, 18, 11, 4, 19, 25, 6]. This consists of using training data to select a kernel out of the family of convex combinations of $p$ base kernels and to learn a predictor based on the kernel selected, these two tasks being performed either in a single stage by solving one optimization as in most studies such as [16], or in subsequent stages as in a recent technique described by [7].

The most frequently used framework for this approach is that of Lanckriet et al. [16], which is both natural and elegant. But, experimental results reported for this method have not shown a significant improvement over the straightforward baseline of training with a uniform combination of base kernels. The more recent two-stage technique for learning kernels presented by Cortes et al. [7] is shown, however, to achieve a better performance than the uniform combination baseline across multiple datasets. The algorithm consists of first learning a non-negative combination of the base kernels using a notion of centered alignment with the target label kernel, and then of using that combined kernel with a kernel-based algorithm to select a hypothesis. Figure 1 illustrates these two learning kernel techniques.

An alternative approach explored by this paper consists of using data to learn a predictor for each base kernel and combine these predictors to define a single predictor, these two tasks being performed in a single stage or in two subsequent stages (see Figure 1). This approach is distinct from the learning kernel one since it does not seek to learn a kernel, however its high-level objective is to address the same problem ($\mathcal{P}$). The predictors returned by this approach are ensembles of kernel predictors (EKPs) or of kernel-based hypotheses.

Note that each of the hypotheses combined belongs to a different set, the reproducing kernel Hilbert space (RKHS) associated to a different kernel. As we shall see later, the hypothesis family of EKPs can contain the one used by learning kernel techniques based on convex combinations of $p$ base kernels. This raises the question of guarantees for learning with the family of hypotheses of EKPs and the comparison of its complexity with that of learning kernels,

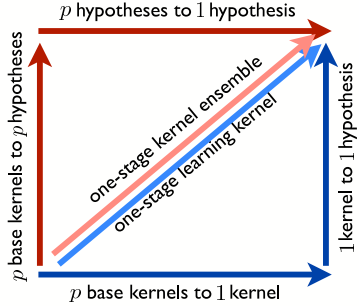

Figure 1: Illustration of different approaches for solving problem ($\mathcal{P}$): learning kernel and ensemble techniques. The path in blue represents the subsequent stages of the two-stage learning kernel algorithm of [7]. Similarly, the path in red represents the two-stage ensemble technique studied here. The standard one-stage technique for learning kernel [16] is represented by the diagonal in light blue and similarly the single-stage EKP technique is indicated by a diagonal in pink.

which we shall address later.

**Relationship with standard ensemble methods** We briefly discuss the connection of the setting examined with that of standard ensemble methods such as boosting. In our setting, an ensemble method is applied to the $p$ hypotheses $h_k$, $k \in [1, p]$, obtained via training in the first stage. The ensemble method we use in our experiments is $L_1$- or $L_2$-regularized linear SVM for a classification task, Lasso or ridge regression for a regression task (augmented with a non-negativity constraint) which enable us to control the norm of the vector of ensemble coefficients with different $L_q$-norms. Of course, for a classification task, other ensemble methods such as boosting could be used instead to combine the hypotheses $h_k$ (without regularization). But, we are not advocating a specific ensemble technique and our analysis is general. As we shall see, the theory we present applies regardless of the specific ensemble method used in the second stage.

Let us point out, however, that the existing margin theory available for ensemble methods [14, 5] will not be very informative in our setting. The existing theory applies to convex combinations of a single hypothesis set $H$. Thus, here, it could apply in two ways: (1) by considering the case where an ensemble method such as boosting is applied to the finite set of base classifiers $H = \{h_1, \ldots, h_p\}$; or (2) by studying the case where $H = \cup_{k=1}^{p} H_k$ is the union of the RKHSs $H_k$ associated to each base kernel $K_k$. In the former case, the learning guarantees for the ensemble classifier would depend on the complexity of the finite set $H$ of hypotheses, which would be of limited interest since this would not directly include any information about the kernels used and since in our setting $h_1, \ldots, h_p$ are not known in advance. In the latter case, the generalization bounds would then be in terms of the complexity of the union $(\cup_{k=1}^{p} H_k)$. Instead, our analysis provides finer learning guarantees in terms of the characteristics of the base kernels $K_k$ defining the Hilbert spaces $H_k$ and the number of kernels $p$, by specifically studying convex regularized non-negative combinations of hypotheses from different spaces. Furthermore, our analysis is given for different $L_q$ regularizations, while the existing bounds are valid only for $L_1$. Finally, note that the application of a boosting algorithm in the second scenario would be very costly since it would require training $p$ kernel-based algorithms at each round.

**Previous work on ensembles of kernel predictors** Ensembles of kernel-based hypotheses have been considered in a number of different contexts and applications of which we name a few. Ideas from standard ensemble techniques of bagging and boosting were used by Kim et al. and other authors [12, 13, 20] to assign weights to SVM hypotheses viewed as base learners, with a linear or non-linear step such as majority vote, least squared error weighting, or a "double-layer hierarchical" method to combine their scores. The authors seem to use the same kernel for training each SVM. SVM ensembles have also been explored to address the problem of training with datasets containing a rare class by repeating the rare training instances across the training sets for individual base classifiers [24]. Finally, learning ensembles with a coupled method by sharing additional parameters between the trained models is studied by [10]. On the theoretical side, leave-one-out and cross-validation bounds were given for kernel-based ensembles by [9], limited to fixed (not learned) combination weights. A recent paper of Koltchinskii and Yuan [15] also studies ensembles of kernel ensembles, but analyzes a rather different form of regularization and deals exclusively with a one-stage algorithm.

**Our contribution** We present both a theoretical and an empirical analysis of EKPs and compare them with several methods for learning kernels, including those of [16] and [7]. We give novel and tight bounds on the Rademacher complexity of the hypothesis sets corresponding to EKPs and compare them with similar recent bounds given by [6] for learning kernels. We show in particular that, while the hypothesis set for EKPs contains that of learning kernels, remarkably, for $L_1$ regularization, the complexity bound for EKPs coincides with the one for learning kernels and thus provides favorable guarantees. We also introduce a natural one-stage learning algorithm for EKPs, analyze its relationship with the two-stage EKP algorithm, and show its close relationship with the algorithm of [16].

Our empirical results include a series of experiments with EKPs based on using $L_1$ and $L_2$ regularization in the second stage for both classification and regression, and a comparison with several algorithms for learning kernels. They demonstrate, in particular, that EKPs achieve a perfor-mance superior to that of learning with a uniform combination of base kernels and that they also typically surpass the one-stage learning kernel algorithm of [16]. EKPs also appear to be competitive against the two-stage kernel learning method of [7] that they outperform in several tasks.

The remainder of this paper is organized as follows. The next section (Section 2) defines the learning scenario for EKPs and the corresponding hypothesis sets. Section 3 presents the results of our theoretical analysis based on the Rademacher complexity of these hypothesis sets. In Section 4, we introduce and discuss a one-stage algorithm for learning EKPs. Section 5 reports the results of our experiments comparing with several algorithms for learning kernels and EKPs on a number of data sets.

## 2 Learning Scenario

This section describes the standard scenario for learning an ensemble of kernel-based hypotheses and introduces much of the notation used in other sections. We denote by $\mathcal{X}$ the input space and by $\mathcal{Y}$ the output space, with $\mathcal{Y} = \{-1, +1\}$ in classification and $\mathcal{Y} \subseteq \mathbb{R}$ in regression.

Let $K_k$ with $k \in [1, p]$ be $p \geq 1$ PDS kernels. We shall denote by $\mathbb{H}_K$ the reproducing kernel Hilbert space (RKHS) associated to a PDS kernel $K$, and by $\|\cdot\|_{\mathbb{H}_K}$ the corresponding norm in that space. In the absence of ambiguity, to simplify the notation, we write $\mathbb{H}_k$ instead of $\mathbb{H}_{K_k}$. In the first stage of the ensemble setting, $p$ hypotheses $h_1, \ldots, h_p$ are obtained by training a kernel-based algorithm using the same sample $S = ((x_1, y_1), \ldots, (x_m, y_m)) \in (\mathcal{X} \times \mathcal{Y})^m$ with each of these kernels. This is typically done using an algorithm based on an optimization of the form $h_k = \operatorname{argmin}_{h \in \mathbb{H}_k} \lambda_k \|h\|^2_{\mathbb{H}_k} + \sum_{i=1}^m L(h(x_i), y_i)$, where $L \colon \mathcal{Y} \times \mathcal{Y} \to \mathbb{R}$ is a loss function convex in its first argument and where $\lambda_k \geq 0$ is a regularization parameter. In our experiments, we use support vector machines (SVMs) [8] in classification tasks and kernel ridge regression (KRR) [21] in regression tasks. These correspond respectively to the hinge loss defined by $L(y, y') = \max(1 - yy', 0)$ and the square loss defined by $L(y, y') = (y' - y)^2$. Since each base hypothesis $h_k$ is learned using a different kernel $K_k$, the regularization parameter $\lambda_k$ obtained by cross-validation is different in each optimization. Equivalently, each base hypothesis $h_k$ is selected from a set $\{h \in \mathbb{H}_k : \|h\|_{\mathbb{H}_k} \leq \Lambda_k\}$ with a distinct $\Lambda_k \geq 0$.

In the second stage, a possibly separate training sample is used to learn a non-negative linear combination of these hypotheses, $\sum_{k=1}^p \mu_k h_k$, with an $L_q$ regularization: $\boldsymbol{\mu} \in \Delta_q$ with $\Delta_q = \{\boldsymbol{\mu} \colon \boldsymbol{\mu} \geq \mathbf{0} \wedge \sum_{k=1}^p \mu_k^q = 1\}$. Thus, the hypothesis set corresponding to such ensembles has the following general form for $L_q$ regularization:

$$\mathcal{E}_p^q = \left\{ \sum_{k=1}^p \mu_k h_k \colon \|h_k\|_{\mathbb{H}_k} \leq \Lambda_k, k \in [1, p], \boldsymbol{\mu} \in \Delta_q \right\}. \quad (1)$$

Our experiments are carried out with an $L_1$ regularization, corresponding to convex combinations of kernels ($q = 1$), or $L_2$ regularization ($q = 2$).

Note that it might be possible to define a tighter hypothesis set describing our learning scenario, in which the weights $\boldsymbol{\mu}$ are further restricted in terms of the first stage solutions $h_k^*$. Since our analysis is meant to be general though and valid for any learning algorithms used in the two stages, it is not clear how this could be achieved. But, in any case, as we shall see in Section 3.1, already with our definition, the learning guarantees for EKPs match the tight learning bounds proven for the learning kernel scenario, which demonstrates favorable guarantees for EKPs.

## 3 Theoretical Analysis

To analyze the complexity of the hypothesis families just defined, we bound, for different values of $q$, their empirical Rademacher complexity $\widehat{\mathfrak{R}}_S(\mathcal{E}_p^q)$ for an arbitrary sample $S$ of size $m$. This immediately yields generalization bounds for EKPs, in particular a margin bound in classification of the form [14, 5]:

$$\forall h \in \mathcal{E}_p^q, \ R(h) \leq \widehat{R}_\rho(h) + \frac{2}{\rho} \widehat{\mathfrak{R}}_S(\mathcal{E}_p^q) + 3\sqrt{\frac{\log \frac{2}{\delta}}{2m}},$$

where $\rho > 0$ is the margin, $\delta > 0$ the confidence level, $R(h)$ the generalization error of $h$, and $\widehat{R}_\rho(h)$ the fraction of the training points with margin less than $\rho$ (i.e. $y_i h(x_i) \leq \rho$). Our proof techniques build on those used by [6] to derive bounds for learning kernels, with which we compare those we obtain for EKPs.

For a sample $S = (x_1, \ldots, x_m)$, the empirical Rademacher complexity of a family of functions $H$ is defined by

$$\widehat{\mathfrak{R}}_S(H) = \frac{1}{m} \operatorname*{E}_{\boldsymbol{\sigma}} \left[ \sup_{h \in H} \sum_{i=1}^m \sigma_i h(x_i) \right],$$

where the expectation is taken over $\boldsymbol{\sigma} = (\sigma_1, \ldots, \sigma_m)^\top$ with $\sigma_i \in \{-1, +1\}$ independent uniform random variables. For any kernel function $K$, we denote by $\mathbf{K} = [K(x_i, x_j)] \in \mathbb{R}^{m \times m}$ its kernel matrix for the sample $S$. The following proposition gives the general form of the Rademacher complexity of the hypothesis set $\mathcal{E}_p^q$.

**Proposition 1.** *Let $q, r \geq 1$ with $\frac{1}{q} + \frac{1}{r} = 1$. For any sample $S$ of size $m$, the empirical Rademacher complexity of the hypothesis set $\mathcal{E}_p^q$ can be expressed as $\widehat{\mathfrak{R}}_S(\mathcal{E}_p^q) = \frac{1}{m} \operatorname{E}_{\boldsymbol{\sigma}} \left[ \|\mathbf{v}_{\boldsymbol{\sigma}}\|_r \right]$ with $\mathbf{v}_{\boldsymbol{\sigma}} = (\Lambda_1 \sqrt{\boldsymbol{\sigma}^\top \mathbf{K}_1 \boldsymbol{\sigma}}, \ldots, \Lambda_p \sqrt{\boldsymbol{\sigma}^\top \mathbf{K}_p \boldsymbol{\sigma}})^\top$.*

*Proof.* By definition of the empirical Rademacher com-

plexity, we can write

$$\widehat{\mathfrak{R}}_S(\mathcal{E}_p^q) = \frac{1}{m}\mathop{\mathrm{E}}_{\boldsymbol{\sigma}}\Big[\sup_{h\in\mathcal{E}_p^q}\sum_{i=1}^m \sigma_i h(x_i)\Big]$$

$$= \frac{1}{m}\mathop{\mathrm{E}}_{\boldsymbol{\sigma}}\Big[\sup_{\substack{\boldsymbol{\mu}\in\Delta_q, h_k\in\mathbb{H}_k \\ \|h_k\|_{\mathbb{H}_k}\leq\Lambda_k}}\sum_{i=1}^m \sigma_i \sum_{k=1}^p \mu_k h_k(x_i)\Big].$$

For any $h_k \in \mathbb{H}_k$, by the reproducing property, for all $x \in \mathcal{X}$, $h_k(x) = \langle h_k, K_k(x,\cdot)\rangle$. Let $\mathbb{H}_{k,S} = \mathrm{span}(\{K_k(x,\cdot)\colon x\in S\})$, then, for $x\in S$, $h_k(x) = \langle h_{k,\|}, K_k(x,\cdot)\rangle$, where $h_{k,\|}$ is the orthogonal projection of $h_k$ over $\mathbb{H}_{k,S}$. Thus, there exist $\alpha_{ki}\in\mathbb{R}, i\in[1,m]$, such that $h_{k,\|} = \sum_{i=1}^m \alpha_{ki}K_k(x_i,\cdot)$. Let $\boldsymbol{\alpha}_k$ denote the vector $(\alpha_{k1},\ldots,\alpha_{km})^\top$, if $\|h_k\|_{\mathbb{H}_k}\leq\Lambda_k$, then

$$\boldsymbol{\alpha}_k^\top\mathbf{K}_k\boldsymbol{\alpha}_k = \|h_{k,\|}\|_{\mathbb{H}_k}^2 \leq \|h_k\|_{\mathbb{H}_k}^2 \leq \Lambda_k^2.$$

Conversely, any $\sum_{i=1}^p \alpha_{ki}K_k(x_i,\cdot)$ with $\boldsymbol{\alpha}_k^\top\mathbf{K}_k\boldsymbol{\alpha}_k\leq\Lambda_k^2$ is the projection of some $h_k\in\mathbb{H}_k$ with $\|h_k\|_{\mathbb{H}_k}^2\leq\Lambda_k^2$. Thus, we can write

$$\widehat{\mathfrak{R}}_S(\mathcal{E}_p^q) = \frac{1}{m}\mathop{\mathrm{E}}_{\boldsymbol{\sigma}}\Big[\sup_{\substack{\boldsymbol{\mu}\in\Delta_q \\ \boldsymbol{\alpha}_k^\top\mathbf{K}_k\boldsymbol{\alpha}_k\leq\Lambda_k^2}}\sum_{k=1}^p \mu_k \sum_{i,j=1}^m \sigma_i\alpha_{kj}K_k(x_i,x_j)\Big]$$

$$= \frac{1}{m}\mathop{\mathrm{E}}_{\boldsymbol{\sigma}}\Big[\sup_{\substack{\boldsymbol{\mu}\in\Delta_q \\ \boldsymbol{\alpha}_k^\top\mathbf{K}_k\boldsymbol{\alpha}_k\leq\Lambda_k^2}}\sum_{k=1}^p \mu_k\boldsymbol{\sigma}^\top\mathbf{K}_k\boldsymbol{\alpha}_k\Big].$$

Fix $\boldsymbol{\mu}$. Since the terms in $\boldsymbol{\alpha}_k$ are not restricted by any shared constraints, they can be optimized independently via

$$\max_{\boldsymbol{\alpha}_k^\top\mathbf{K}_k\boldsymbol{\alpha}_k\leq\Lambda_k^2}\boldsymbol{\sigma}^\top\mathbf{K}_k\boldsymbol{\alpha}_k = \Lambda_k\sqrt{\boldsymbol{\sigma}^\top\mathbf{K}_k\boldsymbol{\sigma}},$$

where we used the fact that by the Cauchy-Schwarz inequality the maximum is reached for $\mathbf{K}^{1/2}\boldsymbol{\sigma}$ and $\mathbf{K}^{1/2}\boldsymbol{\alpha}_k$ collinear. Thus, by the definition of vector $\mathbf{v}_{\boldsymbol{\sigma}}$, we are left with

$$\widehat{\mathfrak{R}}_S(\mathcal{E}_p^q) = \frac{1}{m}\mathop{\mathrm{E}}_{\boldsymbol{\sigma}}\Big[\sup_{\boldsymbol{\mu}\in\Delta_q}\sum_{k=1}^p \mu_k\Lambda_k\sqrt{\boldsymbol{\sigma}^\top\mathbf{K}_k\boldsymbol{\sigma}}\Big]$$

$$= \frac{1}{m}\mathop{\mathrm{E}}_{\boldsymbol{\sigma}}\Big[\sup_{\boldsymbol{\mu}\in\Delta_q}\boldsymbol{\mu}^\top\mathbf{v}_{\boldsymbol{\sigma}}\Big] = \frac{1}{m}\mathop{\mathrm{E}}_{\boldsymbol{\sigma}}[\|\mathbf{v}_{\boldsymbol{\sigma}}\|_r]$$

where the final equality follows from the definition of the dual norm.[1] □

### 3.1 Rademacher complexity of $L_1$-regularized EKPs

**Theorem 1.** *For any sample $S$ of size $m$, the empirical Rademacher complexity of the hypothesis set $\mathcal{E}_p^1$ can be bounded as follows for all integer $r\geq 1$,*

$$\widehat{\mathfrak{R}}_S(\mathcal{E}_p^1) \leq \frac{\sqrt{\eta_0 r\|\mathbf{v}_{\boldsymbol{\Lambda}}\|_r}}{m},$$

---

[1] Note that this proposition differs from the one given by [6] for learning kernels where $\Lambda=1$ and the term $\sqrt{\|\mathbf{u}_{\boldsymbol{\sigma}}\|}$ appears in place of $\|\mathbf{v}_{\boldsymbol{\sigma}}\|$, with $\mathbf{u}_{\boldsymbol{\sigma}}=(\boldsymbol{\sigma}^\top\mathbf{K}_1\boldsymbol{\sigma},\ldots,\boldsymbol{\sigma}^\top\mathbf{K}_p\boldsymbol{\sigma})^\top$.

*where $\mathbf{v}_{\boldsymbol{\Lambda}} = (\Lambda_1^2\mathrm{Tr}[\mathbf{K}_1],\ldots,\Lambda_p^2\mathrm{Tr}[\mathbf{K}_p])^\top$ and $\eta_0=\frac{23}{22}$. Let $\Lambda_\star=\max_{k\in[1,p]}\Lambda_k$. If further $p>1$ and $K_k(x,x)\leq R^2$ for all $x\in\mathcal{X}$ and $k\in[1,p]$, then*

$$\widehat{\mathfrak{R}}_S(\mathcal{E}_p^1) \leq \sqrt{\frac{\eta_0 e\lceil\log p\rceil\Lambda_\star^2 R^2}{m}}.$$

*Proof.* By Proposition 1, $m\widehat{\mathfrak{R}}_S(\mathcal{E}_p^1) = \mathop{\mathrm{E}}_{\boldsymbol{\sigma}}[\|\mathbf{v}_{\boldsymbol{\sigma}}\|_\infty]$, thus

$$m\widehat{\mathfrak{R}}_S(\mathcal{E}_p^1) = \mathop{\mathrm{E}}_{\boldsymbol{\sigma}}\Big[\max_{k\in[1,p]}\Lambda_k\sqrt{\boldsymbol{\sigma}^\top\mathbf{K}_k\boldsymbol{\sigma}}\Big]$$

$$= \mathop{\mathrm{E}}_{\boldsymbol{\sigma}}\Big[\sqrt{\max_{k\in[1,p]}\Lambda_k^2\boldsymbol{\sigma}^\top\mathbf{K}_k\boldsymbol{\sigma}}\Big] = \mathop{\mathrm{E}}_{\boldsymbol{\sigma}}\Big[\sqrt{\|\mathbf{v}'\|_\infty}\Big],$$

with $\mathbf{v}' = (\Lambda_1^2\boldsymbol{\sigma}^\top\mathbf{K}_1\boldsymbol{\sigma},\ldots,\Lambda_p^2\boldsymbol{\sigma}^\top\mathbf{K}_p\boldsymbol{\sigma})^\top$. Since for any $r\geq 1$, $\|\mathbf{v}'\|_\infty\leq\|\mathbf{v}'\|_r$, using Jensen's inequality,

$$m\widehat{\mathfrak{R}}_S(\mathcal{E}_p^1) \leq \mathop{\mathrm{E}}_{\boldsymbol{\sigma}}\Big[\sqrt{\|\mathbf{v}'\|_r}\Big] = \mathop{\mathrm{E}}_{\boldsymbol{\sigma}}\Big[\Big[\sum_{k=1}^p(\Lambda_k^2\boldsymbol{\sigma}^\top\mathbf{K}_k\boldsymbol{\sigma})^r\Big]^{\frac{1}{2r}}\Big]$$

$$\leq \Big[\sum_{k=1}^p\mathop{\mathrm{E}}_{\boldsymbol{\sigma}}\big[(\Lambda_k^2\boldsymbol{\sigma}^\top\mathbf{K}_k\boldsymbol{\sigma})^r\big]\Big]^{\frac{1}{2r}}.$$

The first result then follows the bound $\mathop{\mathrm{E}}_{\boldsymbol{\sigma}}[(\boldsymbol{\sigma}^\top\mathbf{K}\boldsymbol{\sigma})^r] \leq (\eta_0 r\mathrm{Tr}[\mathbf{K}])^r$ which holds by Lemma 1 of [6]. Now, if $K_k(x,x)\leq R^2$ for all $x\in\mathcal{X}$ and $k\in[1,p]$, $\mathrm{Tr}[\mathbf{K}_k]\leq mR^2$ for all $k\in[1,p]$ and

$$\|\mathbf{v}_{\boldsymbol{\Lambda}}\|_r = \Big(\sum_{k=1}^p(\Lambda_k^2\mathrm{Tr}[\mathbf{K}_k])^r\Big)^{1/r} \leq p^{1/r}\Lambda_\star^2 mR^2.$$

Thus, by Theorem 1, for any integer $r>1$, the Rademacher complexity can be bounded as follows

$$\widehat{\mathfrak{R}}_S(\mathcal{E}_p^1) \leq \frac{1}{m}\big[\eta_0 rp^{1/r}\Lambda_\star^2 mR^2\big]^{\frac{1}{2}} = \sqrt{\frac{\eta_0 rp^{\frac{1}{r}}\Lambda_\star^2 R^2}{m}}.$$

The result follows the fact that for $p>1$ $r\mapsto p^{1/r}r$ reaches its minimum at $r_0=\log p$. □

We compare this bound with a similar bound for the hypothesis set based on convex combinations of base kernels used for learning kernels [6], for $\Lambda_1=\ldots=\Lambda_p$:

$$H_p^1 = \Big\{h\in\mathbb{H}_K\colon K=\sum_{k=1}^p \mu_k K_k, \boldsymbol{\mu}\in\Delta_1, \|h\|_{\mathbb{H}_K}\leq\Lambda_\star\Big\}.$$

Remarkably, the theorem shows that the bound on the empirical Rademacher complexity of the hypothesis set for EKPs coincides with the one for $\widehat{\mathfrak{R}}_S(H_p^1)$. It suggests that learning with $\mathcal{E}_p^1$ does not increase the risk of overfitting with respect to learning with $H_p^1$, while offering the opportunity for a smaller empirical error. The theorem also shows that the bound we gave for $\widehat{\mathfrak{R}}_S(\mathcal{E}_p^1)$ is tight since $\mathcal{E}_p^1$ contains $H_p^1$ and since the bound for $\widehat{\mathfrak{R}}_S(H_p^1)$ given by [6] was shown to be tight. The next section examines different $L_q$ regularizations.

## 3.2 Rademacher complexity of $L_q$-regularized EKPs

**Theorem 2.** *Let $q, r \geq 1$ with $\frac{1}{q} + \frac{1}{r} = 1$ and assume that $r$ is an integer. Then, for any sample $S$ of size $m$, the empirical Rademacher complexity of the hypothesis set $\mathcal{E}_p^q$ can be bounded as follows:*

$$\widehat{\mathfrak{R}}_S(\mathcal{E}_p^q) \leq \frac{\sqrt{\eta_0 r} \|\mathbf{u}\|_r}{m},$$

*where $\mathbf{u} = (\Lambda_1 \sqrt{\mathrm{Tr}[\mathbf{K}_1]}, \ldots, \Lambda_p \sqrt{\mathrm{Tr}[\mathbf{K}_p]})^\top$ and $\eta_0 = \frac{23}{22}$. Let $\Lambda_\star = \max_{k \in [1,p]} \Lambda_k$. If further $p > 1$ and $K_k(x,x) \leq R^2$ for all $x \in \mathcal{X}$ and $k \in [1,p]$, then*

$$\widehat{\mathfrak{R}}_S(\mathcal{E}_p^q) \leq \sqrt{\frac{\eta_0 r p^{\frac{2}{r}} \Lambda_\star^2 R^2}{m}}.$$

*Proof.* By Proposition 1 $m\widehat{\mathfrak{R}}(\mathcal{E}_p^q) = \mathrm{E}_{\boldsymbol{\sigma}}[\|\mathbf{v}_{\boldsymbol{\sigma}}\|_r]$. Using this identity and Jensen's inequality gives:

$$m\widehat{\mathfrak{R}}(\mathcal{E}_p^q) = \mathop{\mathrm{E}}_{\boldsymbol{\sigma}}\left[\left(\sum_{k=1}^p (\Lambda_k^2 \boldsymbol{\sigma}^\top \mathbf{K}_k \boldsymbol{\sigma})^{r/2}\right)^{1/r}\right]$$

$$\leq \left(\sum_{k=1}^p \left(\mathop{\mathrm{E}}_{\boldsymbol{\sigma}}\left[(\Lambda_k^2 \boldsymbol{\sigma}^\top \mathbf{K}_k \boldsymbol{\sigma})^r\right]\right)^{1/2}\right)^{1/r}.$$

By the bound $\mathrm{E}_{\boldsymbol{\sigma}}\left[(\boldsymbol{\sigma}^\top \mathbf{K} \boldsymbol{\sigma})^r\right] \leq (\eta_0 r \mathrm{Tr}[\mathbf{K}])^r$, which holds by Lemma 1 of [6],

$$\left(\sum_{k=1}^p \left(\mathop{\mathrm{E}}_{\boldsymbol{\sigma}}\left[(\Lambda_k^2 \boldsymbol{\sigma}^\top \mathbf{K}_k \boldsymbol{\sigma})^r\right]\right)^{1/2}\right)^{1/r}$$

$$\leq \left(\sum_{k=1}^p (\eta_0 r \Lambda_k^2 \mathrm{Tr}[\mathbf{K}_k])^{r/2}\right)^{1/r} = \frac{\sqrt{\eta_0 r}}{m}\|\mathbf{u}\|_r.$$

This proves the first statement. For the second statement, when $\mathrm{Tr}[\mathbf{K}_k] \leq mR^2$ for all $k$, $\|\mathbf{u}\|_r = \left(\sum_{k=1}^p \Lambda_k^r \mathrm{Tr}[\mathbf{K}_k]^{r/2}\right)^{1/r} \leq \left(p^{\frac{2}{r}} \Lambda_\star^2 m R^2\right)^{\frac{1}{2}}$. Thus, in view of the first result, the following holds

$$\widehat{\mathfrak{R}}_S(\mathcal{E}_p^q) \leq \frac{\sqrt{\eta_0 r}}{m} \|\mathbf{u}\|_r \leq \frac{\sqrt{\eta_0 r}}{m} \left((p^{\frac{2}{r}} \Lambda_\star^2 m R^2)^{r/2}\right)^{1/r}$$

$$= \sqrt{\frac{\eta_0 r p^{\frac{2}{r}} \Lambda_\star^2 R^2}{m}}. \qquad \square$$

Here, for $\Lambda_1 = \ldots = \Lambda_p$, the bound on the Rademacher complexity is less favorable than the one for learning kernels with the similar family:

$$H_p^q = \Big\{h \in \mathbb{H}_K : K_{\boldsymbol{\mu}} = \sum_{k=1}^p \mu_k K_k, \boldsymbol{\mu} \in \Delta_q, \|h\|_{\mathbb{H}_K} \leq \Lambda_\star\Big\}.$$

The bound given by [6] for $\mathfrak{R}_S(H_p^q)$ is smaller exactly by a factor of $p^{1/(2r)}$. Thus, as an example, here, for $L_2$ regularization, the guarantee for learning with EKPs is less favorable by a factor of $\sqrt{p}$, which, for large $p$, can be significant.

## 4 Single-Stage Learning Algorithm

This section introduces and discusses a single-stage learning algorithm for EKPs, which turns out to be closely related to a standard algorithm for learning kernels. The natural framework for learning EKPs consists of the two stages detailed in Section 2 where $p$ hypotheses $h_k$ are learned using different kernels in the first stage and a mixture weight $\boldsymbol{\mu}$ is learned in the second stage to combine them linearly.

Alternatively, one can consider, as for learning kernels [16], a single-stage learning algorithm for EKPs. For a fixed $\boldsymbol{\mu} \in \Delta_q$, define $\mathcal{H}_{\boldsymbol{\mu}}$ by $\mathcal{H}_{\boldsymbol{\mu}} = \{\sum_{k=1}^p \mu_k h_k : h_k \in \mathbb{H}_k, k \in [1,p]\}$. A hypothesis $h$ may admit different expansions $\sum_{k=1}^p \mu_k h_k$ (even for a fixed $\boldsymbol{\mu}$), thus we denote by $\overline{\mathcal{H}}_{\boldsymbol{\mu}}$ the multiset of all hypotheses with their different expansions and denote by $h_1, \ldots, h_p$ the corresponding base hypotheses. A natural algorithm for a single-stage ensemble learning is thus one which penalizes the empirical loss of the final hypothesis $h = \sum_{k=1}^p \mu_k h_k(x)$, while controlling the norm of each base hypothesis $h_k$. The following is the corresponding optimization problem:

$$\min_{\boldsymbol{\mu} \in \Delta_q} \min_{h \in \overline{\mathcal{H}}_{\boldsymbol{\mu}}} \sum_{i=1}^m L(h(x_i), y_i)$$

$$\text{subject to: } \|h_k\| \leq \Lambda_k, k \in [1,p].$$

Introducing Lagrange variables $\lambda_k \geq 0, k \in [1,p]$, this can be equivalently written as

$$\min_{\boldsymbol{\mu} \in \Delta_q} \min_{h \in \overline{\mathcal{H}}_{\boldsymbol{\mu}}} \sum_{k=1}^p \lambda_k \|h_k\|_{K_k}^2 + \sum_{i=1}^m L(h(x_i), y_i). \quad (2)$$

**Relationship with two-stage algorithm.** Note that, in the case $q=1$, by the convexity of the loss function with respect to its first argument, for any $i \in [1,m]$, $L(h(x_i), y_i) \leq \sum_{k=1}^p \mu_k L(h_k(x_i), y_i)$. If we replace the empirical loss in (2) with this upper bound, we obtain:

$$\min_{\boldsymbol{\mu} \in \Delta_1} \min_{h \in \overline{\mathcal{H}}_{\boldsymbol{\mu}}} \sum_{k=1}^p \lambda_k \|h_k\|_{K_k}^2 + \sum_{k=1}^p \mu_k \sum_{i=1}^m L(h_k(x_i), y_i).$$

In this optimization, for a fixed $\boldsymbol{\mu}$, the terms depending on each $k \in [1,p]$ are decoupled and can be optimized independently. Thus, proceeding in this way precisely coincides with the two-stage ensemble learning algorithm as described in Section 2.

**Relationship with one-stage learning kernel algorithm.** The main algorithmic framework for learning kernels in a single-stage is based on the following optimization problem:

$$\min_{\boldsymbol{\mu} \in \Delta_q} \min_{h \in \mathbb{H}_{K_{\boldsymbol{\mu}}}} \lambda \|h\|_{K_{\boldsymbol{\mu}}}^2 + \sum_{i=1}^m L(h(x_i), y_i), \quad (3)$$

where $\mathbb{H}_{K_{\boldsymbol{\mu}}}$ is the RKHS associated to the PDS kernel $K_{\boldsymbol{\mu}} = \sum_{k=1}^{p} \mu_k K_k$, $\lambda \geq 0$ is a regularization parameter, and $q = 1$ [16] or $q = 2$. We shall compare the algorithms based on the optimizations (2) and (3). Our proof will make use of the following general lemma.

**Lemma 1.** *Let $K$ be a PDS kernel. For any $\lambda > 0$, $\mathbb{H}_{\lambda K} = \mathbb{H}_K$ and $\langle \cdot, \cdot \rangle_{\lambda K} = \frac{1}{\lambda} \langle \cdot, \cdot \rangle_K$, in particular $\|\cdot\|_{\lambda K}^2 = \frac{1}{\lambda} \|\cdot\|_K^2$.*

*Proof.* It is clear that $\mathbb{H}_{\lambda K} = \mathbb{H}_K$ since elements of $\mathbb{H}_{\lambda K}$ can be obtained from $\mathbb{H}_K$ bijectively by multiplication by $\lambda$. Now, for any $h \in \mathbb{H}_{\lambda K} = \mathbb{H}_K$, by the reproducing property, for all $x \in \mathcal{X}$,

$$h(x) = \langle h, K(x, \cdot) \rangle_K$$

and $\quad h(x) = \langle h, \lambda K(x, \cdot) \rangle_{\lambda K} = \lambda \langle h, K(x, \cdot) \rangle_{\lambda K} .$

Matching these equalities shows that for all $h$, $\langle h, K(x, \cdot) \rangle_K = \lambda \langle h, K(x, \cdot) \rangle_{\lambda K}$. Thus, for all $h' = \sum_{i \in I} \alpha_i K(x_i, \cdot)$, $\langle h, h' \rangle_K = \lambda \sum_{i \in I} \alpha_i \langle h, K(x, \cdot) \rangle_{\lambda K} = \lambda \langle h, h' \rangle_{\lambda K}$. This shows that $\langle \cdot, \cdot \rangle_K = \lambda \langle \cdot, \cdot \rangle_{\lambda K}$ and concludes the proof of the lemma. □

**Proposition 2.** *For $\lambda_k = \lambda \mu_k$ for all $k \in [1, p]$, the optimization problem for learning EKPs (2) and the one for learning kernels (3) are equivalent.*

*Proof.* Fix $\boldsymbol{\mu} \in \Delta_q$.

$$\min_{h \in \overline{\mathcal{H}}_{\boldsymbol{\mu}}} \sum_{k=1}^{p} \lambda_k \|h_k\|_{K_k}^2 + \sum_{i=1}^{m} L(h(x_i), y_i)$$
$$= \min_{h \in \mathcal{H}_{\boldsymbol{\mu}}} \min_{\substack{h = \sum_{k=1}^{p} \mu_k h_k \\ h_k \in \mathbb{H}_k}} \left\{ \sum_{k=1}^{p} \lambda_k \|h_k\|_{K_k}^2 \right\} + \sum_{i=1}^{m} L(h(x_i), y_i)$$
$$= \min_{h \in \mathcal{H}_{\boldsymbol{\mu}}} \min_{\substack{h = \sum_{k=1}^{p} h'_k \\ h'_k \in \mathbb{H}_k}} \left\{ \sum_{k=1}^{p} \frac{\lambda_k}{\mu_k^2} \|h'_k\|_{K_k}^2 \right\} + \sum_{i=1}^{m} L(h(x_i), y_i)$$

(replacing $\mu_k h_k$ with $h'_k$)

$$= \min_{h \in \mathcal{H}_{\boldsymbol{\mu}}} \lambda \min_{\substack{h = \sum_{k=1}^{p} h'_k \\ h'_k \in \mathbb{H}_k}} \left\{ \sum_{k=1}^{p} \frac{1}{\mu_k} \|h'_k\|_{K_k}^2 \right\} + \sum_{i=1}^{m} L(h(x_i), y_i)$$

(assumption on $\lambda_k$s)

$$= \min_{h \in \mathcal{H}_{\boldsymbol{\mu}}} \lambda \min_{\substack{h = \sum_{k=1}^{p} h'_k \\ h'_k \in \mathbb{H}_{K'_k}}} \left\{ \sum_{k=1}^{p} \|h'_k\|_{K'_k}^2 \right\} + \sum_{i=1}^{m} L(h(x_i), y_i)$$

(Lemma 1),

with $K'_k = \mu_k K_k$. By a theorem of Aronszajn (Theorem p.353 [3]), if $h = \sum_{k=1}^{p} h_k$, with $h_k \in \mathbb{H}_{K'_k}$, then $h \in \mathbb{H}_K$ and $\min_{h = \sum_{k=1}^{p} h'_k, h'_k \in \mathbb{H}_{K'_k}} \{\sum_{k=1}^{p} \|h_k\|_{K'_k}^2\} = \|h\|_K^2$ with $K = \sum_{k=1}^{p} K'_k$. Thus,

$$\min_{h \in \overline{\mathcal{H}}_{\boldsymbol{\mu}}} \sum_{k=1}^{p} \lambda_k \|h_k\|_{K_k}^2 + \sum_{i=1}^{m} L(h(x_i), y_i)$$
$$= \min_{h \in \mathbb{H}_K} \lambda \|h\|_K^2 + \sum_{i=1}^{m} L(h(x_i), y_i).$$

Taking the minimum over $\boldsymbol{\mu} \in \Delta_q$ yields the statement of the proposition. □

Thus, under the assumptions of the proposition, the one-stage algorithm for EKPs returns exactly the same solution as the one for learning kernels. A similar result was given by [15] for a Lasso-type regularization using a lemma of [18]. In general, however, this one-stage algorithm for EKPs is not practical for large values of $p$ since the number of parameters $\lambda_k$ to determine simultaneously using cross-validation becomes too large. In view of this drawback, we did not use this algorithm in our experiments.

## 5 Experiments

We did a series of experiments with EKPs and compared their performance with that of several existing learning kernel methods across several datasets from the UCI, UCSD-MKL and Delve repositories for both the classification and the regression setting.

We experimented both with $L_1$-regularized ensembles (denoted **L1-ens**) and $L_2$-regularized ensembles (**L2-ens**). For the first stage, the base hypotheses were obtained by using SVMs for classification or KRR for regression. In the second stage, for $L_1$-regularized ensembles, $L_1$-regularized SVM was used for classification, Lasso in regression. In the case of $L_2$ regularization, standard SVM and KRR were used in the second stage. In all cases for the second stage, the primal version of the problem was solved with a linear kernel over the predictions of the hypotheses of the first stage and is augmented with an explicit constraint $\boldsymbol{\mu} \geq \mathbf{0}$. The ensemble performance was compared to that of the single combination kernel selected by the following algorithms, used in conjunction with SVM or KRR.

**unif:** kernel-based algorithm with a uniform kernel combination, $\mathbf{K}_{\boldsymbol{\mu}} = \sum_{k=1}^{p} \mu_k \mathbf{K}_k = \frac{\Lambda}{p} \sum_{k=1}^{p} \mathbf{K}_k$.

**os-svm:** one-stage kernel learning method that selects an $L_1$-regularized non-negative weighted kernel combination for SVM [16]. The following is the corresponding optimization problem:

$$\min_{\boldsymbol{\mu}} \max_{\boldsymbol{\alpha}} \ 2\boldsymbol{\alpha}^\top \mathbf{1} - \boldsymbol{\alpha}^\top \mathbf{Y}^\top \mathbf{K}_{\boldsymbol{\mu}} \mathbf{Y} \boldsymbol{\alpha}$$

subject to: $\boldsymbol{\mu} \geq \mathbf{0}, \text{Tr}[\mathbf{K}_{\boldsymbol{\mu}}] \leq \Lambda, \boldsymbol{\alpha}^\top \mathbf{y} = 0, \mathbf{0} \leq \boldsymbol{\alpha} \leq \mathbf{C}$.

**os-krr:** one-stage kernel learning method that selects an $L_2$-regularized non-negative weighted kernel combination

## CLASSIFICATION

|    | $\gamma_1, \gamma_p, p$ | $N$ | **unif** | **os-svm** | **align** | **alignf** | **L1-ens** | **L2-ens** |
|----|---|---|---|---|---|---|---|---|
| G  | $-4, 3, 8$ | 1000 | 25.9±1.8 | 26.0±2.6 | 25.8±2.9 | 24.7±2.1 | 25.4±1.5 | 25.3±1.4 |
| PA | $\cdot, \cdot, 10$ | 694 | 8.9±2.6 | 8.5±2.7 | 8.4±2.8 | 9.7±1.9 | 7.1±3.0 | 7.2±3.0 |
| PB | $\cdot, \cdot, 10$ | 694 | 10.0±1.7 | 9.3±2.4 | 9.4±1.9 | 9.3±1.8 | 9.7±2.5 | 8.1±1.5 |
| SM | $-12, -7, 6$ | 1000 | 18.7±2.8 | 20.9±2.8 | 18.5±2.3 | 18.7±2.5 | 15.4±1.3 | 15.7±1.7 |
| SM | $-12, -7, 6$ | 2000 | 15.7±2.8 | 18.4±2.6 | 16.1±3.0 | 16.0±1.2 | 13.7±1.1 | 13.8±1.0 |
| SM | $-12, -7, 6$ | 4601 | 12.3±0.9 | 13.9±0.9 | 12.4±0.9 | 13.1±1.0 | 9.4±0.5 | 9.8±0.6 |

## REGRESSION

|   | $\gamma_1, \gamma_p, p$ | $N$ | **unif** | **os-krr** | **align** | **alignf** | **L1-ens** | **L2-ens** |
|---|---|---|---|---|---|---|---|---|
| I | $-3, 3, 7$ | 351 | .467±.085 | .457±.085 | .467±.093 | .446±.093 | .437±.086 | .433±.084 |
| K | $-3, 3, 7$ | 1000 | .138±.005 | .137±.005 | .136±.005 | .129±.01 | .120±.005 | .120±.005 |

Table 1: Performance of several kernel combination algorithms across both regression and classification datasets: german (G), protein fold class-7 vs. all (PA) and class-16 vs. all (PB), spambase (SM), ionosphere (I) and kinematics (K). Average misclassification error is reported for classification, average RMSE for regression, and in both cases one standard deviation as measured across 5 trials.

for KRR. The following is the corresponding optimization problem:

$$\min_{\substack{\boldsymbol{\mu} \geq 0 \\ \|\boldsymbol{\mu}\|_2 \leq \Lambda}} \max_{\boldsymbol{\alpha}} -\lambda \boldsymbol{\alpha}^\top \boldsymbol{\alpha} - \boldsymbol{\alpha}^\top \mathbf{K}_{\boldsymbol{\mu}} \boldsymbol{\alpha} + 2\boldsymbol{\alpha}^\top \mathbf{y}$$

**align:** two-stage $L_1$-regularized alignment-based technique presented by [7] which weights each base kernel proportionally to the alignment $\mu_k \propto \frac{\langle \mathbf{K}_k, \mathbf{yy}^\top \rangle_F}{\|\mathbf{K}_k\|_F}$, where $\langle \cdot, \cdot \rangle_F$, denotes the Frobenius product, of the centered kernel matrix $\mathbf{K}_k$ and the kernel matrix of the training labels $\mathbf{yy}^\top$, resulting in a combination kernel $\mathbf{K}_{\boldsymbol{\mu}} = \sum_{k=1}^p \mu_k \mathbf{K}_k$ with $\sum_{k=1}^p \mu_k \leq \Lambda$.

**alignf:** another two-stage $L_1$-regularized technique of [7], jointly maximizing the alignment of the kernel matrix with the target labels kernel taking in to account the correlation between kernel matrices:

$$\mathbf{K}_{\boldsymbol{\mu}} = \operatorname*{argmax}_{\mathbf{K}_{\boldsymbol{\mu}}, \frac{\mu}{\Lambda} \in \Delta_1} \frac{\langle \mathbf{K}_{\boldsymbol{\mu}}, \mathbf{yy}^\top \rangle_F}{\|\mathbf{K}_{\boldsymbol{\mu}}\|_F}.$$

We note that, for **align** and **alignf**, using $L_2$-regularization only scales the $L_1$-regularized solution by a factor that can be absorbed into $\Lambda$. Thus, this difference in regularization would provide no practical difference in performance.

The experimental setup is modeled after that of [7]. For each dataset, several Gaussian kernels of the form $K(x, x') = \exp(-\gamma \|x - x'\|^2)$, with different bandwidth parameters $\gamma$, are used as base kernels. The set of $\gamma$s used are $\{2^{\gamma_1}, 2^{\gamma_1+1}, \ldots, 2^{\gamma_p}\}$, where $\gamma_1$ and $\gamma_p$ and the number of resulting kernels $p$ are indicated in Table 1 for each dataset. In case of the protein fold dataset, the kernels provided by the UCSD-MKL repository are used. The norm of the combination weights is controlled by the parameter $\|\boldsymbol{\mu}\|_q \leq \Lambda$, for either $q=1$ or $q=2$ as appropriate. This parameter is selected based on the best average performance on a validation set. The regularization parameter of KRR ($\lambda$) or SVM ($C$) is held constant since it is effectively only the ratio $\Lambda/\lambda$ or $\Lambda/C$ that determines the solution.

The average error and standard deviation reported is for 5-fold cross validation using a total of $N$ data-points, where three folds are used for training, one fold for validation, and one fold for measuring the test error. That is, the training set size $m = \frac{3}{5}N$. For the two stage methods, the training set is further split into two independent training sets. The first one is used to train the base hypotheses and the second one to learn the mixture weights. The ratio of the split, chosen from the set $\{10/90, 20/80, \ldots, 90/10\}$, is decided by the best average performance on the validation set.

Table 1 shows that, in several datasets, the performance of the EKP algorithms is superior to that of the uniform kernel baseline **unif**, which has proven to be difficult to improve upon in the past in the learning kernel literature. EKPs also achieve a better performance than the standard one-stage learning kernel algorithms, **os-svm** or **os-krr**, in several datasets. Finally, we observe that EKPs also improve upon the alignment based methods, which had previously reported the best performance among learning kernel techniques [7]. This improvement is substantial for some data sets, e.g. spambase data sets.[2] These improvements over the best learning kernel results reported are remarkable and very encouraging for further studies of EKPs.

If given access to only a single CPU, the time it takes to train the EKPs can be substantially longer than any of the other methods we used since $p$ hypotheses must be trained as opposed to a single one. For the spambase dataset with 1,200 training points and using an Intel Xeon 2.33GHz processor with 16GB of total memory, training the 6 base hypotheses sequentially and learning the best combination takes about 1.3 minutes, while the other compared approaches can be trained within 20 seconds. However, if the number of base hypotheses is reasonable and a distributed system is used, as is the case in our experiments, the base hypotheses can be trained on different processors, which results in a clock time similar to that of other methods.

---

[2] Our empirical results somewhat differ from those of [7] for some of the same data sets. This is most likely because we use a split training set in order to match the setting of EKPs. However, even comparing to the results of [7], the improvement of EKP is still significant.

## 6 Conclusion

We presented a general analysis of learning with ensembles of kernel predictors, including a theoretical analysis based on the Rademacher complexity of the corresponding hypothesis sets, the study of a natural one-stage algorithm and its connection with a standard algorithm used for learning kernels, and the results of extensive experiments in several tasks. Our empirical results show that their performance is often significantly superior to the straightforward use of a uniform combination of kernels for learning, which has been difficult to improve upon using algorithms for learning kernels. They also suggest that EKPs can outperform, sometimes substantially, even the best existing algorithms recently reported for learning kernels.